\documentclass{article}
\usepackage{amsmath,epsfig}
\usepackage[preprint]{spconfa4}
\usepackage[font=scriptsize,labelfont=bf]{caption}
\usepackage[belowskip=-10pt,aboveskip=3pt]{caption}

\usepackage{multirow}
\usepackage{tabularx}
\usepackage{dsfont}
\usepackage{url}
\usepackage[tight]{subfigure}
\usepackage{color}
\usepackage{subfigure}
\usepackage{amsthm}
\usepackage[export]{adjustbox}
\usepackage{comment}

\usepackage[utf8]{inputenc} 
\usepackage[T1]{fontenc}    
\usepackage{url}            
\usepackage{booktabs}       
\usepackage{amsfonts}       
\usepackage{nicefrac}       
\usepackage{microtype}      
\usepackage{pdfpages}
\usepackage{amssymb}

\def\argmax{\operatornamewithlimits{arg\,max}}

\definecolor{dg}{rgb}{0,0.694,0.298}
\definecolor{purple}{rgb}{0.4,0.176,0.569}
\definecolor{Remblue}{rgb}{0.368,0.745,0.941}

\usepackage{pifont}
\usepackage{xspace}
\makeatletter
\DeclareRobustCommand\onedot{\futurelet\@let@token\@onedot}
\def\@onedot{\ifx\@let@token.\else.\null\fi\xspace}
\def\eg{\emph{e.g}\onedot} 
\def\ie{\emph{i.e}\onedot} 
 
\def\etc{\emph{etc}\onedot}

\makeatother

\definecolor{citecolor}{RGB}{65,105,225}
\usepackage[breaklinks=true,colorlinks,citecolor=citecolor,bookmarks=false]{hyperref}

\copyrightnotice{Copyright notice – 978-1-6654-3864-3/21/\$31.00~\copyright 2021 IEEE}

\let\OLDthebibliography\thebibliography
\renewcommand\thebibliography[1]{
  \OLDthebibliography{#1}
  \setlength{\parskip}{0pt}
  \setlength{\itemsep}{0pt plus 0.3ex}
}

\begin{document}\sloppy

\topmargin=0mm 
\def\x{{\mathbf x}}
\def\L{{\cal L}}

\title{DeepMix: Online Auto Data Augmentation for Robust\\Visual Object Tracking}

%
\name{Ziyi Cheng$^{1^{\ast}}$, Xuhong Ren$^{2^{\ast}}$, Felix Juefei-Xu$^{3}$, Wanli Xue$^{2^{\dagger}}$, Qing Guo$^{4\dagger}$, Lei Ma$^{5,6,1}$, Jianjun Zhao$^{1}$\thanks{$^{\ast}$Ziyi Cheng and Xuhong Ren are co-first authors and contribute equally to this work. ${^{\dagger}}$Wanli Xue and Qing Guo are corresponding authors (xuewanli@email.tjut.edu.cn and tsingqguo@ieee.org).}}
\address{$^{1}$Kyushu University, Japan \quad
$^{2}$Tianjin University of Technology, China \\
$^{3}$Alibaba Group, USA \quad
$^{4}$Nanyang Technological University, Singapore\\
$^{5}$University of Alberta, Canada \quad $^{6}$Alberta Machine Intelligence Institute (Amii), Canada
}

\maketitle

\begin{abstract}
Online updating of the object model via samples from historical frames is of great importance for accurate visual object tracking. 
Recent works mainly focus on constructing effective and efficient updating methods while neglecting the training samples for learning discriminative object models, which is also a key part of a learning problem.
In this paper, we propose the \textit{DeepMix} that takes historical samples' embeddings as input and generates augmented embeddings online, enhancing the state-of-the-art online learning methods for visual object tracking. 
More specifically, we first propose the \textit{online data augmentation} for tracking that online augments the historical samples through object-aware filtering. Then, we propose \textit{MixNet} which is an offline trained network for performing online data augmentation within one-step, enhancing the tracking accuracy while preserving high speeds of the state-of-the-art online learning methods.
The extensive experiments on three different tracking frameworks, \ie, DiMP, DSiam, and SiamRPN++, and three large-scale and challenging datasets, \ie, OTB-2015, LaSOT, and VOT, demonstrate the effectiveness and advantages of the proposed method.

%
\end{abstract}
\begin{keywords}
data augmentation, online updating, visual object tracking, deepmix, mixnet
\end{keywords}

\section{Introduction}\label{sec:intro}


Visual object tracking (VOT) is one of the most widely studied computer vision approaches that can produce the trajectory of the moving object from a sequence of frames. It has seen ubiquitous applications ranging from navigation for robots, intelligent video surveillance, smart logistics, robotics for manufacturing, \etc. Based on the frame processing procedure, the VOT can be divided into online tracking and offline tracking. For online tracking, only the current frame and the previous frames can be used to determine the tracking result for the current frame, and the tracking results for the previous frames, once computed, can no longer be altered based on later frames. For offline tracking, the tracking results can be produced after having access to all the frames. Needless to say, the settings of online tracking make it more suitable for real-world applications and deployment.

\begin{figure}[t]
\centering
\includegraphics[height=2.9in,width=1.0\linewidth]{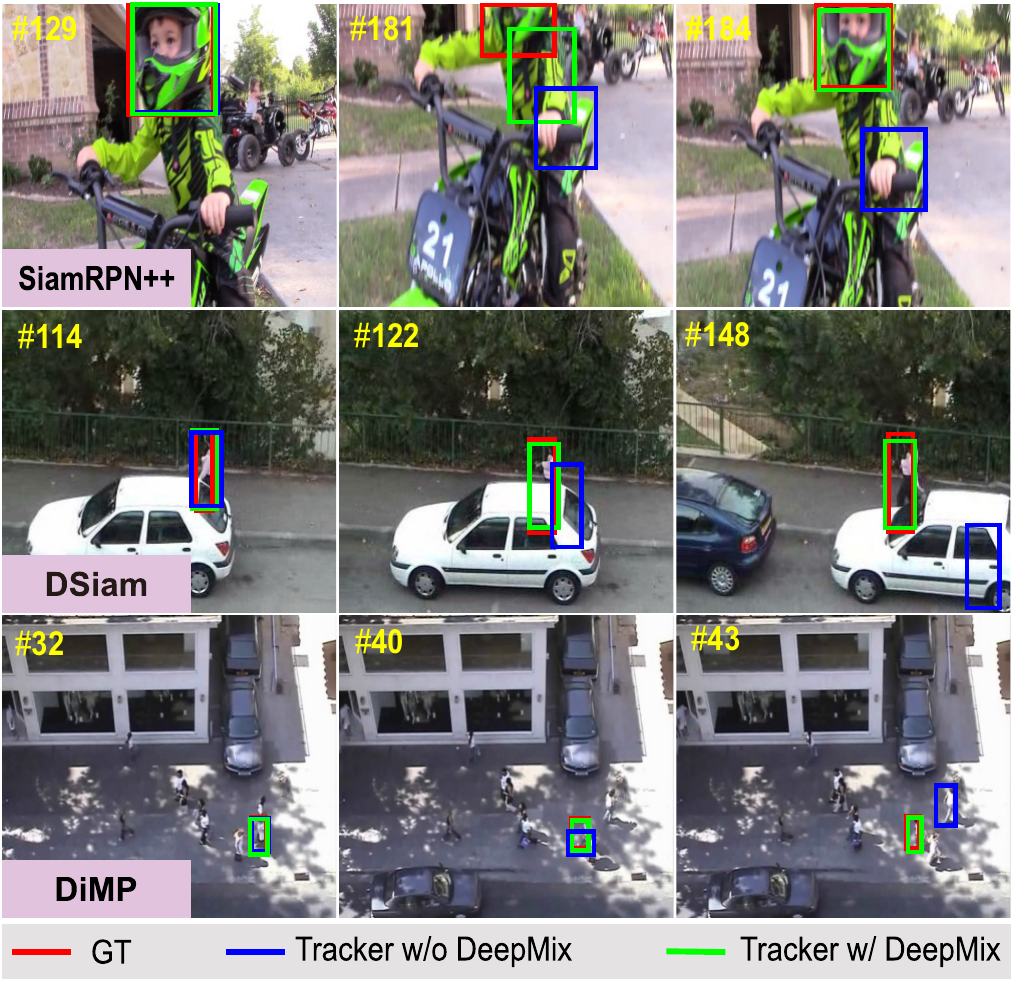}
\caption{Examples of three state-of-the-art trackers, \eg, SiamRPN++ \cite{li2019siamrpn++}, DSiam \cite{guo2017learning,guo2020spark}, and DiMP \cite{Bhat2019ICCV} with or without the proposed \textit{DeepMix}.}
\label{fig:teaser}
\end{figure}

In the early days, VOT is implemented using correlation filters (CFs) with notable works such as the minimum output sum of squared error (MOSSE) CF \cite{bolme2010visual}, kernelized correlation filter (KCF) \cite{henriques2014high}, spatially regularized discriminative correlation filter \cite{danelljan2015learning,zhou2017selective,zhang2018fast,feng2019dynamic,guo2020selective}, \etc. With the fast-paced development of deep learning approaches that are catered towards computer vision problems, deep learning based VOT has gained tremendous popularity. Methods such as SiamFC \cite{Bertinetto16-2} and subsequent SiamRPN \cite{li2018high}, DiMP \cite{BhatDGT19ICCV}, DSiam \cite{guo2017learning}, DaSiamRPN \cite{zhu2018distractor}, SiamRPN++ \cite{Li2019CVPR}, \etc, have progressively produced better results on various VOT benchmarks \cite{Wu15, kristan2018sixth, Fan2019LaSOT,guo2021exploring}. We think that online updating of the object model via samples from historical frames is of great importance for accurate visual object tracking. These mentioned recent works mainly focus on constructing powerful deep backbones or effective and efficient online updating methods while neglecting the training samples for learning discriminative object models, which is also a key part 
of a learning problem in our opinion. 


In this paper, we propose the \textit{DeepMix} that takes historical samples' embeddings as input and generates augmented embeddings online, enhancing the state-of-the-art online learning methods for visual object tracking. 
More specifically, we first propose the \textit{online data augmentation} for tracking that online augments the historical samples through object-aware filtering. Then, we propose \textit{MixNet} which is an offline trained network for performing online data augmentation within one-step, enhancing the tracking accuracy while preserving high speeds of the state-of-the-art online learning methods. MixNet predicts different convolution parameters dramatically for object and background regions, respectively, according to the input training samples, thus is able to generate effective training samples for online data augmentation. 
As shown in Fig.~\ref{fig:teaser}, our \textit{DeepMix} let three state-of-the-art trackers, \ie, DiMP \cite{BhatDGT19ICCV}, DSiam \cite{guo2017learning, guo2020spark}, and SiamRPN++ \cite{li2019siamrpn++}, localize objects more accurately.
The extensive experiments on the above three different tracking frameworks and three large-scale and challenging datasets, \ie, OTB-2015, LaSOT, and VOT, further demonstrate the effectiveness and advantages of the proposed method.


\section{Related Work}\label{sec:related}

\subsection{General Visual Object Tracking}
In visual object tracking task, Siamese network-based methods \cite{Bertinetto16-2,guo2017learning,zhu2018distractor,Li2019CVPR,chen2020siamese} have been a popular solution. Amony them, SiamFC \cite{Bertinetto16-2} is an initial implementation, which extracts the deep features from template and search region and performs cross-correlation to predict object position. SiamRPN \cite{li2018high} proposes to add a bounding box regressor to the Siamese Network as done in the object detection task. 

Another effective solution treats tracking as a online learning problem. They \cite{guo2016structure,BhatDGT19ICCV,danelljan2020probabilistic,wang2020tracking,bhat2020know} train a classifier from past frames with an online update strategy and distinguish the target from background via the classifier. DiMP \cite{BhatDGT19ICCV} applies meta-learning formulation by collecting historical frame feature representation to classify objects. ROAM \cite{yang2020roam} offline trains an LSTM \cite{hochreiter1997long} to generate the adaptive learning rate to enhance online training. 
Several works also focus on the training sample management during the online tracking. \cite{Danelljan16} manages samples by assigning different weights to training samples.
ECO \cite{Danelljan16ECO} employs a Gaussian mixture model (GMM) to choose more distinguishing samples.
UpdateNet \cite{zhang2019learning} uses history templates to generating a more robust template through a CNN which is similar to our DeepMix. However, it aims to update effective templates.
Although above works have shown great advantages of managing training samples, they ignore how to make more effective use of online samples, \eg, data augmentation. In this work, we focus on the data augmentation for the state-of-the-art trackers during tracking.

\subsection{Data Augmentation Methods}
Data augmentation is an important method to improve generalization performance of deep models. 
Early works \cite{ciregan2012multi,krizhevsky2017imagenet,guo2017calibration} employ cropping, horizontal and vertical flips, and rotation to generate diverse data. Recently, AutoAugment \cite{cubuk2018autoaugment} automatically searches for augmentation policies given a predefined set of transformations, which needs a great quantity of training time. Several studies reduce the search costs significantly, such as Population-based augmentation (PBA) \cite{ho2019population} and fast AutoAugment (FAA) \cite{lim2019fast}. These works focus on augmenting a single image.

Recent works \cite{hendrycks2019augmix,zhang2017mixup,yun2019cutmix} propose to mix multiple images for data augmentation, inspiring our research on mixing multiple training samples for tracking.
Specifically, Mixup \cite{zhang2017mixup} utilizes an element-wise combination of two images. CutMix \cite{yun2019cutmix} replaces a portion of an image with the contents of another image. AugMix \cite{hendrycks2019augmix} merges multiple images where each image is processed by several augment operations randomly, making the trained model see diverse samples. All of these methods focus on image-level augmentation and do not consider the speed for online data. In contrast to previous works, we propose the DeepMix allowing efficient online data augmentation for visual object tracking.





\section{Method}\label{sec:method}
In this section, we first discuss the background and motivation of this work and formulate the online data augmentation for tracking in  Sec.~\ref{subsec:background} and \ref{subsec:onlineaug}. Then, we propose the \textit{MixNet} in Sec.~\ref{subsec:MixNetAug} to realize effective and efficient online data augmentation for tracking. Finally, we detail how to embed MixNet into state-of-the-art trackers (\ie, SiamRPN++ \cite{Li2019CVPR}, DSiam \cite{guo2017learning,guo2020spark}, and DiMP \cite{BhatDGT19ICCV}) in Sec.~\ref{subsec:sota_mixnet}.

\subsection{Background and Motivation} \label{subsec:background}
Given a live video $\mathcal{V}=\{\mathbf{I}_t\}_{t=1}^{T}$ having $T$ frames and the object bounding box annotated at the first frame (\ie, $\mathbf{b}_1$) , we aim to estimate  
the object's position and size at the subsequent $T-1$ frames. Most of the state-of-the-art methods complete this task by maintaining an object model for matching it with the subsequent frames. In general, we formulate the object localization at $t$-th frame as  
\begin{align}\label{eq:loc}
\mathbf{p}_t=\argmax_{\mathbf{p}}{\mathbf{M}_t}[\mathbf{p}]= \argmax_{\mathbf{p}}{\text{loc}(\varphi(\mathbf{I}_t),\theta_t)},
\end{align}
where the $\mathbf{M}_t$ is a heat map whose the maximum (\ie, ${\mathbf{M}_t}[\mathbf{p}_t]$) indicates the object's position in the frame $\mathbf{I}_t$, and it can be calculated by the $\varphi(\mathbf{I}_t)$ and $\theta_t$
where $\varphi(\cdot)$ is the backbone network for extracting embedding.

The object model $\theta_t$ determines the localization accuracy, which is initialized at the first frame and updated at subsequent frames. For example, in the popular Siamese network-based trackers \cite{Bertinetto16-2,Li2019CVPR}, the object model is constructed by using the embedding of the object at the first frame, \ie, $\theta_t =\varphi(\mathbf{I}_1)$, and the localization is implemented by using cross-correlation, \ie, $\text{loc}(\varphi(\mathbf{I}_t),\theta_t)=\varphi(\mathbf{I}_t)*\varphi(\mathbf{I}_1)$.
More recently, \cite{BhatDGT19ICCV} proposes DiMP that uses an online updated classifier for localization (\ie, the $\text{loc}(\cdot)$ is set as a convolution layer).

The state-of-the-art trackers (\eg, DiMP \cite{BhatDGT19ICCV} and DSiam \cite{guo2017learning}) online update the object model to adapt object and background appearance variation. In general, we formulate the object model updating via
%
\begin{align}\label{eq:update}
\theta_{t+1}=\text{update}(\theta_t,\mathcal{X}_t)
\end{align}
%
where $\mathcal{X}_t$ denotes the set of training samples that are cropped from the historical frames in which the objects are previously detected. For example, DSiam \cite{guo2017learning} proposes to online update the object model of the Siamese network via a transformation that is learned from the previous frame. 
DiMP \cite{BhatDGT19ICCV} updates the classifier's parameters through a pre-trained model predictor that takes 50 samples from previous frames as inputs.

Note that, the updating process is a typical learning module and recent works have demonstrated that data augmentation is of great importance for enhancing the accuracy of image classification under various interferences \cite{hendrycks2020augmix}.  
Following similar ideas, we aim to explore how to online augment effective training samples, \ie, $\mathcal{X}_t$, for the state-of-the-art trackers with existing updating methods, \ie, online transformation in DSiam \cite{guo2017learning} and model predictor in DiMP \cite{BhatDGT19ICCV}.

%
\begin{figure}[t]
\centering
\includegraphics[width=0.8\linewidth]{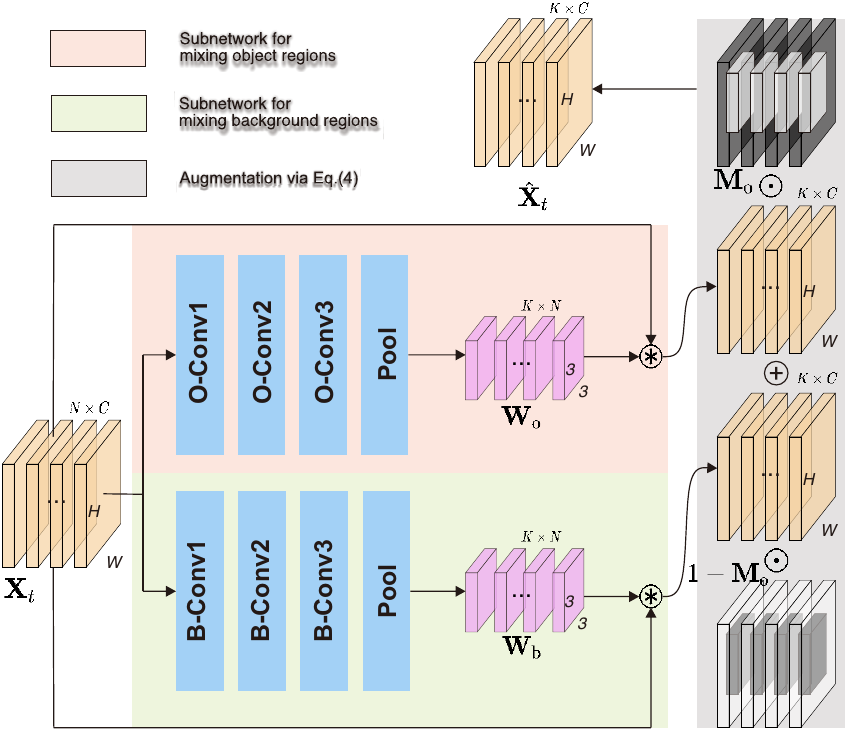}
\caption{Architecture of MixNet that contains two subnetworks to estimate filtering parameters for mixing object and background regions, respectively.}
\label{fig:arch}
\end{figure}
%

\subsection{Online Data Augmentation for Tracking} \label{subsec:onlineaug}

A simple way for online data augmentation is to borrow the existing techniques and conduct augmentation on the collected historical training samples, \ie, $\mathcal{X}_t$, through
%
\begin{align}\label{eq:aug}
\hat{\mathcal{X}}_{t}=\text{aug}(\mathcal{X}_t, \mathcal{T})
\end{align}
%
where $\mathcal{T}$ denotes the set of sample-level transformations (\eg, adding noise, blur, and rain) and $\hat{\mathcal{X}}_t$ is the augmented samples.
The state-of-the-art data augmentation techniques are usually employed in the offline training process with random and diverse degradation-related sample-level transformations \cite{hendrycks2020augmix}. 
For example, AugMix \cite{hendrycks2020augmix} conducts multiple random augmentations for a raw sample and mixes them up for training the image classifiers. 
However, these methods could not adapt to the online data augmentation for visual object tracking directly due to the following reasons: \ding{182} the random sample-level transformations would generate new samples that require high time costs to extract their deep features, slowing down the trackers significantly. \ding{183} the transformations are based on degradation factors (\eg, noise, blur, fog, rain, \etc) that can corrupt the raw samples, leading to the less discriminative object model. 

To address above challenges, we propose the online data augmentation on embeddings of training samples. Specifically, we first extract embeddings of training samples in $\mathcal{X}_t$ and get $\{\varphi(\mathbf{I}_i)\in\mathds{R}^{C\times W \times H}|\mathbf{I}_i\in\mathcal{X}_t\}$.
Then, we concatenate all embeddings and obtain a tensor $\mathbf{X}_t\in\mathds{R}^{N\times C\times W \times H}$ where $N$ is the number of training samples in $\mathcal{X}_t$. 
Our goal is to map the tensor $\mathbf{X}_t$ to a new counterpart denoted as $\hat{\mathbf{X}}_t\in\mathds{R}^{K\times C\times W \times H}$ that can be fed into existing updating modules to produce a more effective object model. 
Note that, performing augmentation on the embedding level is much more efficient than that on the sample level, which alleviates the first challenge.
In terms of the second challenge, we mix embeddings of all samples with the guidance of previously localization results. 
Intuitively, the interested object might be at any position in the scene during the video capturing process and it is reasonable to augment the training samples by putting the object to possible background regions.
To this end, given $\mathbf{X}_t$ and the detected bounding boxes $\{\mathbf{b}_i\in\mathds{R}^{4\times 1}|i=1,\ldots,N\}$ of $N$ training samples, we can split the samples to object and background regions and mixing them up to produce new samples. 
We formulate this process by
%
\begin{align}\label{eq:mix}
\hat{\mathbf{X}}_t=(\mathbf{W}_\text{o}\circledast \mathbf{X}_t)\odot\mathbf{M}_\text{o}+(\mathbf{W}_\text{b}\circledast \mathbf{X}_t)\odot(1-\mathbf{M}_\text{o}),
\end{align}
%
where $\mathbf{M}_\text{o}\in\mathds{R}^{N\times C\times W\times H}$ are binary masks for the $N$ training samples where the elements within the object regions are set to one while others are zero. The object regions are obtained according to the detection results $\{\mathbf{b}_i\in\mathds{R}^{4\times 1}|i=1,...,N\}$ and the `$\odot$' denotes the element-wise multiplication. 
Besides, the `$\circledast$' denotes the convolution layer while the tensor $\mathbf{X}_t$ is filtered by $\mathbf{W}_{\{\text{o or b}\}}\in\mathds{R}^{K\times N\times 3\times 3}$ with the padding hyper-parameter to be one. The variable $K$ denotes the number of samples in the output tensor.
More specifically, for the $c$th channel of $\mathbf{X}_t$ (\ie, $\mathbf{X}_t[c]\in\mathds{R}^{N\times W\times H}$), we filter it with $\mathbf{W}_{\{\text{o or b}\}}$ and get the $c$th channel of $\hat{\mathbf{X}}_t[c]\in\mathds{R}^{K\times W\times H}$.
%
%
%
Intuitively, $\mathbf{W}_{\{\text{o or b}\}}$ indicates how to fuse the $N$ training samples and get $K$ new samples. The $\mathbf{W}_{\text{o}}$ and $\mathbf{W}_{\text{b}}$ take the charge of mixing object and background regions, respectively.

However, to make above method work, we should consider the following issues: \ding{182} how to estimate $\mathbf{W}_{\text{o}}$ and $\mathbf{W}_{\text{ b}}$ to fit different cases? \ding{183} how to make the above module to be efficient? To alleviate these issues, we propose the \textit{MixNet} that is able to produce augmented data in one step.

\subsection{MixNet for Efficient Online Data Augmentation}\label{subsec:MixNetAug}
We propose \textit{MixNet} that takes the $\mathbf{X}_t$ as the input and predict the kernels $\mathbf{W}_{\text{o}}$ and $\mathbf{W}_{\text{ b}}$ that are suitable for $\mathbf{X}_t$. We can use the pre-trained \textit{MixNet} to generate the kernels in a one-step way, leading to efficient online data augmentation. We show the architecture of \textit{MixNet} in Fig.~\ref{fig:arch}. 
Specifically, MixNet contains two sub-networks for predicting the $\mathbf{W}_{\text{o}}$ and $\mathbf{W}_{\text{ b}}$, respectively. 
The two sub-networks share the same architecture but have independent parameters.
The architecture has three convolution layers with a kernel size of $3\times 3$ and an averaging pooling layer.
Note that, we can embed MixNet into diverse tracking frameworks by properly setting the input and output channels for them.

\begin{table}[t]
\footnotesize
\centering
\caption{Comparison with SOTA Trackers under OPE setup.} \label{tab:otb&lasot}
\begin{tabular}{lcccc}
  \toprule
  Dataset & \multicolumn{2}{c}{OTB-2015} & \multicolumn{2}{c}{ LaSOT}\\
   
  \midrule
    Metrics  & AUC &Prec &Success &Prec \\
  \midrule
 
  MLT \cite{choi2019deep}   &  0.611&- &  0.368&-   \\
  GradNet  \cite{li2019gradnet}  &  0.639&0.861 &  0.365 & 0.351   \\
  ATOM \cite{Martin2019CVPR}      &  0.667&0.879 &  0.514&0.505  \\
  SiamDW \cite{choi2019deep}    & 0.674& - &  0.384&-  \\
  POST \cite{wang2020post} & 0.678&0.907 &  0.481&0.463 \\
  CRPN \cite{Fan2019CVPR}  &  0.675&- &  0.455&-  \\
  ASRCF \cite{Dai2019CVPR} &  0.692&0.922 & 0.359&0.337 \\
   MAML \cite{wang2020tracking}      &  0.712&- & 0.523&-  \\
   \midrule
  DSiam     & 0.646 & 0.845 &  0.438 & 0.431 \\
  DSiam-DeepMix       &\textbf{ 0.658} &\textbf{ 0.861 } & \textbf{ 0.439} &\textbf{ 0.431} \\
  \midrule
  SiamRPN++ &0.650  &0.853 &  0.447 &0.446 \\
  SiamRPN++-DeepMix  & \textbf{0.663 }& \textbf{0.870} & \textbf{ 0.459} & \textbf{0.463}\\

  \midrule
  DiMP    & 0.660 & 0.859 & 0.532 & 0.532  \\
  DiMP-DeepMix       & \textbf{0.683} & \textbf{0.890} &  \textbf{0.536}   & \textbf{0.538}\\
  \bottomrule
\end{tabular} \vspace{-5pt}
\end{table} 

\subsection{Implementation for SOTA Trackers}\label{subsec:sota_mixnet}

In this part, we detail the way of using our method for three state-of-the-art trackers, \ie, SiamRPN++ \cite{li2018high,li2019siamrpn++}, DSiam \cite{guo2017learning,guo2020spark}, and DiMP \cite{BhatDGT19ICCV}.
Simply, we can embed \textit{DeepMix} into these trackers by transforming their training samples and get $\hat{\mathbf{X}}_t$. Then, we mix it with the raw training samples (\ie, $\mathbf{X}_t$) by $\alpha_1\hat{\mathbf{X}}_t+\alpha_2\mathbf{X}_t$) and feed it into the updating modules. For all examples, we fix $\alpha_1=0.05$ and $\alpha_2=0.8$ and discuss its influence in Sec.~\ref{subsec:ablation}.

\textbf{DSiam and SiamRPN++ with DeepMix.} We collect historical samples ($\mathbf{X}_t$ size is 15$\times$256$\times$29$\times$29) to generate the kernel ($\mathbf{W}_{\{\text{o or b}\}}$ size is 15$\times$1$\times$3$\times$3) and then filter with the features (its size is 1$\times$256$\times$29$\times$29) of the current frame. Finally, DeepMix output the new samples ($\hat{\mathbf{X}}_t$ size is 1$\times$256$\times$29$\times$29).  

\textbf{DiMP with DeepMix.} DiMP stores samples to train the classifier, we directly input these samples ($\mathbf{X}_t$ size is 50$\times$256$\times$22$\times$22) and obtain new samples with the same size as the Fig.~\ref{fig:arch} illustrate.
We directly train the MixNet along with its embedded trackers. It means that we can simply use the original training program and training data of the targeted trackers to train their own MixNets.
We make some minor modifications to MixNet to adapt to different trackers. 

\textbf{Training details}. For DSiam and SiamRPN++, the original training program uses a pair of images (\ie, template and search region) as a training sample. For each sample, 
%
%
We apply data augmentation strategies on a template to construct a training set containing 15 samples. We input them to MixNet and generate a new sample to mix with the original template.
We implement the SGD optimizer with a weight decay of 0.0005, base lr of 0.005, and momentum of 0.9. We train the MixNet for 40 epochs and 6000 samples per epoch.

In terms of DiMP, its training program pick up 3 images from each video as training samples for its model predictor. In order to match our MixNet during testing, we change it to 50 images from each video for MixNet. We apply SGD optimizer with weight decay of 0.0005 for all parameter layers, base lr of 0.005, and momentum of 0.9.We train the MixNet for 50 epochs and 1000 videos per epoch.
%


\begin{table}[t]
\footnotesize
\centering
\caption{Comparison with SOTA Trackers on VOT2018} \label{tab:vot}
\begin{tabular}{lccc}
  \toprule
    Metrics  &  EAO &   Accuracy& Robustness     \\
  \midrule
  DaSiamRPN \cite{zhu2018distractor} &  0.383 &  0.586 &  0.276  \\
  SiamMask \cite{Wang2019CVPR}   &  0.387 & 0.642 & 0.295  \\  
   MAML \cite{wang2020tracking}      &  0.392 &  0.635 &  0.220  \\
  UpdateNet \cite{zhang2019learning} &  0.393 & - & -  \\
  ATOM \cite{Martin2019CVPR}      &  0.401 &  0.590 &  0.204  \\
   SiamDW \cite{Zhang2019CVPR}    &  0.405 &  0.597 &  0.234  \\
  \midrule
  DSiam     & 0.266 & 0.577 &  0.421 \\
  DSiam-DeepMix      & \textbf{0.287 }&\textbf{ 0.58 } & \textbf{ 0.407 }\\
 
  \midrule
  SiamRPN++ & 0.348 & 0.583 &  0.29  \\
  SiamRPN++-DeepMix  &\textbf{ 0.405 }& \textbf{0.597} & \textbf{ 0.234} \\
  
  \midrule
  DiMP      & 0.214 & 0.578 &  0.553 \\
  DiMP-DeepMix        & \textbf{0.234} &\textbf{ 0.612} & \textbf{ 0.51 } \\
  \bottomrule
\end{tabular} \vspace{-5pt}
\end{table}

\section{Experiments}\label{sec:exp}

\subsection{Setups}

\textbf{Datasets, metrics, and baseline.} 
We evaluate DeepMix on three tracking benchmarks: VOT-2018 \cite{kristan2018sixth} (60 videos, 356 frames average length), LaSOT \cite{Fan2019LaSOT} (280 videos, 2448 frames average length), OTB-2015 \cite{Wu15} (100 videos, 590 frames average length).
VOT-2018 implements a reset-based evaluation that once the object is lost, the tracker is restarted with the ground truth box five frames later and gets a penalty. The main evaluation criterion is expected average overlap (EAO) \cite{kristan2015visual}. The higher EAO indicates better performance.
OTB-2015 and LaSOT only give the trackers the ground-truth of initial frame and obtain bounding box sequence, terms \textit{one-pass evaluation}. We use the AUC, which represents the area under the curve of the success plot, to evaluate the performance of trackers. 

We compare DeepMix-based trackers with six top methods on VOT2018, \ie, DaSiamRPN \cite{zhu2018distractor}, SiamMask \cite{Wang2019CVPR}, MAML \cite{wang2020tracking}, UpdateNet \cite{zhang2019learning}, ATOM \cite{Martin2019CVPR}, SiamDW \cite{Zhang2019CVPR}. We also choose five excellent trackers on OTB2015, \ie, MLT \cite{choi2019deep}, GradNet \cite{li2019gradnet}, POST \cite{wang2020post}, CRPN \cite{Fan2019CVPR}, ASRCF \cite{Dai2019CVPR}.

\subsection{State-of-the-art Comparison} 
We compare DeepMix with the state-of-the-art methods on three challenging tracking benchmarks. Respectively, the backbones of DiMP, DSiam and SiamRPN++ are ResNet18 \cite{he2016identity}, AlexNet \cite{krizhevsky2017imagenet} and MobileV2 \cite{sandler2018mobilenetv2} on account of DeepMix's extreme improvement for a simple network.

\textbf{OTB-2015 and LaSOT}. We report results on the OTB-2015 and LaSOT datasets in Table \ref{tab:otb&lasot}. Results show that: \ding{182} DeepMix improves all of its original trackers. \ding{183} DeepMix has a significant improvement of 2.3 percentage points of AUC and raises the ranking of DiMP from fifth to third compared with other trackers on OTB-2015. DiMP-DeepMix also achieves the top success score on LaSOT. \ding{184} DiMP collects historical samples' embeddings and trains the predictor online. Our MixNet is designed for data augmentation, thus DeepMix has better compatibility with DiMP and achieves more improvement than DSiam and SiamRPN++.

\textbf{VOT2018}. We report results in Table \ref{tab:vot}. DeepMix with three trackers still is in effect. 
Unlike testing on other datasets, DiMP collects 250 samples for online training when testing on VOT2018 and costs extreme memory with DeepMix. Therefore, we still implement the same hyperparameters as evaluating on OTB-2015. Although it has achieved much lower score than its report, the results also prove the effectiveness of DeepMix. SiamRPN-DeepMix obtains a striking 0.057 improvement on EAO. Even though it uses MobileV2 as the backbone, achieves state-of-the-art on VOT2018.

\begin{table}[t]
\footnotesize
\centering
\caption{Ablation analysis of different backbones} \label{tab:ablation back}
\begin{tabular}{lccc}
  \toprule
   Metrics   &  AUC &   Precision	 &  NormPrecision     \\
    
  \midrule
  DiMP~(ResNet18)  & 0.660  &  0.859  &   0.807     \\
  DiMP~(ResNet18)-DeepMix  &\textbf{ 0.683 } & \textbf{ 0.890 }  &   \textbf{  0.834 }   \\
  \midrule
  DiMP~(ResNet50)   & 0.684  & 0.894 &  0.842     \\
  DiMP~(ResNet50)-DeepMix  & \textbf{0.694}  &   \textbf{0.900 }   & \textbf{ 0.852 }    \\
  
  
  \bottomrule
\end{tabular}\vspace{-15pt}
\end{table}

\subsection{Ablation study}\label{subsec:ablation}

\textbf{DeepMix with different backbones.} In order to validate the generality of DeepMix with different backbones, we present the result on OTB-2015 dataset in Table \ref{tab:ablation back}. 
It shows that: \ding{182} DeepMix can take stable effect for any network architecture. \ding{183} DeepMix has more improvement on ResNet18-based than ResNet50-based model. It probably because more powerful networks are less dependent on DeepMix.


\textbf{Validation of MixNet.} 
We implement a naive data augmentation method as the baseline to validate the effectiveness of the MixNet.
That is, we calculate the filtering parameters, \ie, $\mathbf{W}_\text{o}$ and $\mathbf{W}_\text{b}$, by online optimizing an objective function via the gradient descent to replace the proposed MixNet. 
Specifically, we define an objective function that is the $L_2$ distance between the predicted heat map and a Gaussian map having the highest score on the detected position.
Then, at $t$th frame, we can minimize the objective function by tuning the $\mathbf{W}_\text{o}$ and $\mathbf{W}_\text{b}$ via the gradient descent for ten iterations.
We denote this method as `DeepMix-Opt' and compare it with the final version DeepMix based on the DiMP tracker and OTB-2015 dataset.
As shown in Table \ref{tab:ablation ONOFF}, DeepMix-Opt via online iterative optimization can also enhance the tracking accuracy but immensely increase the computational cost, slowing down the DiMP significantly. 

As Fig.~\ref{fig:arch} shows, MixNet has two branches and outputs two filters ($\mathbf{W}_\text{o}$ and $\mathbf{W}_\text{b}$). We test another version of MixNet: keep only one branch and output one convolution kernel, then filters with samples $\mathcal{X}_t$, regardless of object or background. We term it as DeepMix-single. 
As shown in Table \ref{tab:ablation ONOFF}, DeepMix-single outperforms DiMP with a competitive speed, it still is weaker than the final version DeepMix. Therefore, learning different patterns of objects or backgrounds is an important method for online data augmentation.
In contrast, DeepMix with the MixNet achieves much higher accuracy improvement with only 1 FPS speed decrease, demonstrating the effectiveness and advantages of the proposed MixNet.

\begin{table}[t]
\footnotesize
\centering
\caption{Comparing three variants of our method, \ie, DeepMix-Opt,DeepMix-single and DeepMix, on DiMP tracker and OTB-2015 to validate the effectiveness of MixNet.} \label{tab:ablation ONOFF}
\begin{tabular}{lcccc}
  \toprule
     Metrics &  AUC &   Prec. &  NormPrec.  & FPS  \\
  \midrule
  DiMP  &   0.660  &  0.859  &   0.807  & 27.0 \\   
   \midrule
  DiMP-DeepMix-Opt   &\textbf{0.667} &\textbf{0.873 }& \textbf{0.816} &\textbf{10} \\
  DiMP-DeepMix-single &\textbf{0.676} &\textbf{0.884 }&\textbf{0.831}&\textbf{26.5}\\
  \midrule
  DiMP-DeepMix   &\textbf{0.683}&\textbf{0.890}&\textbf{0.834} &\textbf{26.0} \\
  \bottomrule
\end{tabular} \vspace{-15pt}
\end{table} 
%

\section{Conclusion}\label{sec:concl}


In this work, we have taken a deep dive into the data augmentation aspect for improving online visual object tracking, a long-overlooked facet in this domain. Specifically, we have proposed the DeepMix as a complete pipeline that takes historical samples' embeddings as input and generates augmented online, thus enhancing the state-of-the-art online learning methods for visual object tracking. To this end, we have proposed the \textit{online data augmentation} for tracking that online augments the historical samples through object-aware filtering. Then, we have further proposed the \textit{MixNet} which is an offline trained deep neural network for performing online data augmentation within one-step, for boosting the tracking accuracy while preserving high speeds of the state-of-the-art online learning methods. We have carried out extensive experiments on three different tracking frameworks, \ie, DiMP, DSiam, and SiamRPN++, and on three large-scale and challenging datasets, \ie, OTB-2015, LaSOT, and VOT. The experimental results have demonstrated and verified the effectiveness and advantages of the proposed method. 

In the future, we plan to further study the online data augmentation for visual object tracking by considering different degradation, \eg, motion blur \cite{guo2020watch}, rain \cite{zhai2020s,guo2021efficientderain}, illumination variation \cite{fu2021auto, tian2020bias}, \etc.


\small 
\textbf{Acknowledgement.} This work was supported in part by the National Natural Science Foundation of China under Grant 61906135, 62020106004, and 92048301, Tianjin Science and Technology Plan Project under Grant 20JCQNJC01350, JSPS KAKENHI Grant No.20H04168, 19K24348, 19H04086, and JST-Mirai Program Grant No.JPMJMI18BB and JPMJMI20B8, Japan, and Natural Science Foundation of Tianjin under Grant KJZ40420200017. This work was also supported by the Canada CIFAR AI program.

\scriptsize
\bibliographystyle{IEEEbib}
\bibliography{ref_short}

\end{document}